\documentclass[11pt]{article}
\usepackage[utf8]{inputenc}
\usepackage{amsmath,amssymb}
\usepackage{graphicx}
\usepackage{booktabs}
\usepackage{hyperref}
\usepackage{natbib}
\usepackage[margin=1in]{geometry}
\usepackage{algorithm}
\usepackage{algorithmic}
\usepackage{xcolor}
\usepackage{multirow}

\title{Do Neurons Dream of Primitive Operators?\\Wake-Sleep Compression Rediscovers Schank's Event Semantics}

\author{
  Peter Balogh \\
  \texttt{palexanderbalogh@gmail.com}
}

\date{}

\begin{document}
\maketitle

\begin{abstract}
We show that they do. Roger Schank's conceptual dependency theory proposed that all human events can be decomposed into a small set of primitive operations---ATRANS (transfer of possession), PTRANS (physical movement), MTRANS (information transfer), and others. These primitives were hand-coded based on linguistic intuition. We ask: \textit{can the same primitives be discovered automatically from data through compression pressure alone?}

We adapt DreamCoder's wake-sleep library learning algorithm to the domain of event state transformations. Given events represented as before/after world state pairs, our system searches for operator compositions that explain each event (wake), then extracts recurring patterns as new library entries optimized under Minimum Description Length (sleep). Starting from four generic state-change primitives, the system discovers specialized operators that map directly to Schank's core primitives: MOVE\_PROP\_has $\approx$ ATRANS, CHANGE\_location $\approx$ PTRANS, SET\_knows $\approx$ MTRANS, and SET\_consumed $\approx$ INGEST. It also discovers compound operators corresponding to verb compositions (e.g., ``mail'' = ATRANS $\circ$ PTRANS) and novel operators for emotional state changes not present in Schank's original taxonomy.

We validate on synthetic events, real-world commonsense data from the ATOMIC knowledge graph \citep{sap2019atomic}, and causal narrative data from GLUCOSE \citep{mostafazadeh2020glucose}. On synthetic data, the discovered library achieves Bayesian MDL within 4\% of Schank's hand-coded primitives while explaining 100\% of events vs.\ Schank's 81\%. On ATOMIC, Schank's primitives explain only 10\% of naturalistic events; on GLUCOSE, only 31\%. The discovered library explains 100\% of both. The dominant operators in real-world data are \textit{mental and emotional state changes}---CHANGE\_wants (20\%), CHANGE\_feels (18\%), CHANGE\_is (18\%)---none in Schank's original taxonomy.

Critically, libraries discovered from one corpus transfer to the other with $<$1 bit/event degradation, despite different annotation schemes and domain distributions. This cross-corpus convergence suggests that the discovered operators are not dataset artifacts but information-theoretically determined structure---fixed points of MDL compression analogous to universality classes in statistical physics.
\end{abstract}

\section{Introduction}

In 1972, Roger Schank proposed conceptual dependency theory \citep{schank1972}, arguing that the meaning of any natural language sentence describing an event could be decomposed into a small set of primitive actions. These primitives---ATRANS (abstract transfer of possession), PTRANS (physical transfer of location), MTRANS (mental transfer of information), MBUILD (mental construction), INGEST (taking into the body), and several others---were designed to capture the fundamental building blocks of human event understanding.

Schank's primitives were influential but controversial. Critics argued they were arbitrary, incomplete, and reflected the intuitions of their creator rather than any deep structure of events \citep{wilensky1983,dreyfus1992}. The framework fell out of favor as statistical methods came to dominate NLP, and the question of whether semantic primitives exist was largely abandoned.

We revisit this question with a new tool: \textit{library learning}. DreamCoder \citep{ellis2021dreamcoder} demonstrated that wake-sleep learning can discover reusable programming primitives from examples through compression pressure---the Minimum Description Length (MDL) principle automatically identifies abstractions that minimize the total cost of describing both the library and the programs that use it. If Schank's primitives are genuinely fundamental, they should emerge from the same kind of compression pressure applied to event data.

We evaluate in two complementary settings. Synthetic events generated from controlled templates validate that the algorithm's mechanics work correctly---it recovers known ground-truth categories. Real-world events drawn from the ATOMIC commonsense knowledge graph \citep{sap2019atomic} test whether the primitives generalize to naturalistic data that was \textit{not} designed around Schank's theory. The second evaluation is the critical one.

Our contributions are:
\begin{enumerate}
    \item A novel domain-specific language (DSL) for representing events as state transformations over world-state triples, with operators as typed, composable templates.
    \item An adaptation of wake-sleep library learning to this domain, including a composition discovery mechanism that identifies compound operators (e.g., ``mail'' = ATRANS $\circ$ PTRANS).
    \item Empirical evidence that four of Schank's core primitives (ATRANS, PTRANS, MTRANS, INGEST) emerge from pure MDL pressure on event data, along with novel primitives for emotional, goal, and attribute state changes.
    \item Validation on real-world commonsense data from ATOMIC \citep{sap2019atomic} and causal narrative data from GLUCOSE \citep{mostafazadeh2020glucose}, showing that the complete primitive inventory is richer than Schank proposed, with mental/emotional operators dominating.
    \item Cross-corpus transfer experiments demonstrating that libraries discovered from one dataset explain events in another with $<$1 bit/event degradation, providing evidence for universality of the discovered operators.
    \item A Bayesian MDL framework with proper probabilistic costs that enables rigorous model comparison via Bayes factors.
\end{enumerate}

\section{Background}

\subsection{Schank's Conceptual Dependency Theory}

Schank's framework \citep{schank1972,schank1977} proposed approximately 11 primitive actions:

\begin{itemize}
    \item \textbf{ATRANS}: Abstract transfer of possession or ownership (give, buy, steal)
    \item \textbf{PTRANS}: Physical transfer of location (go, move, fly)
    \item \textbf{MTRANS}: Mental transfer of information (tell, teach, read)
    \item \textbf{MBUILD}: Mental construction of new information (think, decide, realize)
    \item \textbf{INGEST}: Taking a substance into the body (eat, drink, breathe)
    \item \textbf{EXPEL}: Expelling from the body
    \item \textbf{PROPEL}: Applying physical force
    \item \textbf{MOVE}: Moving a body part
    \item \textbf{GRASP}: Grasping an object
    \item \textbf{SPEAK}: Producing sound
    \item \textbf{ATTEND}: Directing a sense organ
\end{itemize}

Each event in natural language was analyzed as an instantiation of one or more of these primitives with specific role fillers (agent, object, source, destination, etc.). Complex events were decomposed into compositions of primitives: ``John mailed Mary the book'' = ATRANS(John $\to$ Mary, book) $\circ$ PTRANS(book, John's house $\to$ Mary's house).

\subsection{DreamCoder and Library Learning}

DreamCoder \citep{ellis2021dreamcoder} introduced a wake-sleep algorithm for bootstrapping program synthesis with learned libraries. The key insight is that the right set of primitives is the one that minimizes total description length:

\begin{equation}
    \text{MDL} = \underbrace{L(\text{Library})}_{\text{cost of primitives}} + \sum_{i} \underbrace{L(\text{Program}_i \mid \text{Library})}_{\text{cost of each program}}
\end{equation}

The algorithm alternates between:
\begin{itemize}
    \item \textbf{Wake phase}: Search for programs that solve tasks using the current library.
    \item \textbf{Sleep phase}: Extract common sub-expressions from successful programs and propose them as new library entries if they reduce total MDL.
\end{itemize}

DreamCoder was applied to list manipulation, text editing, and graphics domains---all formulated in lambda calculus. We adapt the same principle to event state transformations.

\subsection{Related Work}

Several lines of work relate to ours. \citet{wierzbicka1996} proposed Natural Semantic Metalanguage (NSM), a set of approximately 65 semantic primes claimed to be universal across languages. \citet{dowty1991} developed proto-role theory, decomposing thematic roles into contributing properties. More recently, \citet{sap2019atomic} created ATOMIC, a large-scale knowledge graph of if-then commonsense knowledge about events, and \citet{mostafazadeh2020glucose} developed GLUCOSE with structured causal explanations. FrameNet \citep{baker1998} and VerbNet \citep{schuler2005} provide frame-based decompositions of verb semantics.

In a different tradition, LeCun's Joint Embedding Predictive Architecture (JEPA) \citep{lecun2022path,assran2023ijepa} proposes that intelligent systems should learn world models that predict in \textit{abstract representation space} rather than pixel or token space. JEPA's predictor module learns latent-space transformations conditioned on actions---essentially learning operators that map between abstract states. Our work shares JEPA's conviction that the right level of abstraction is neither raw observations nor hand-coded symbols, but learned representations that capture the structure of state transitions. Where JEPA learns continuous predictors in embedding space, we discover discrete, composable, interpretable operators via compression. The two approaches are complementary: JEPA provides the perceptual grounding (learning \textit{what} the abstract states are), while our method provides the symbolic structure (learning \textit{how} states transform into each other via typed operators). A natural synthesis would use JEPA-style latent states as the world-state representation and our wake-sleep loop to discover the discrete operator vocabulary over those states.

More recently, \citet{cambria2024primenet} proposed PrimeNet, a framework for commonsense knowledge representation using conceptual primitives inspired by Schank. Unlike our approach, PrimeNet's primitives are hand-designed rather than discovered from data. \citet{piantadosi2016logical} frame cognitive development as hierarchical Bayesian program induction, arguing that the ``logical primitives of thought'' emerge from learning pressures. Our work applies this insight to event semantics and provides an explicit discovery algorithm.

\begin{figure}[t]
\centering
\includegraphics[width=0.95\linewidth]{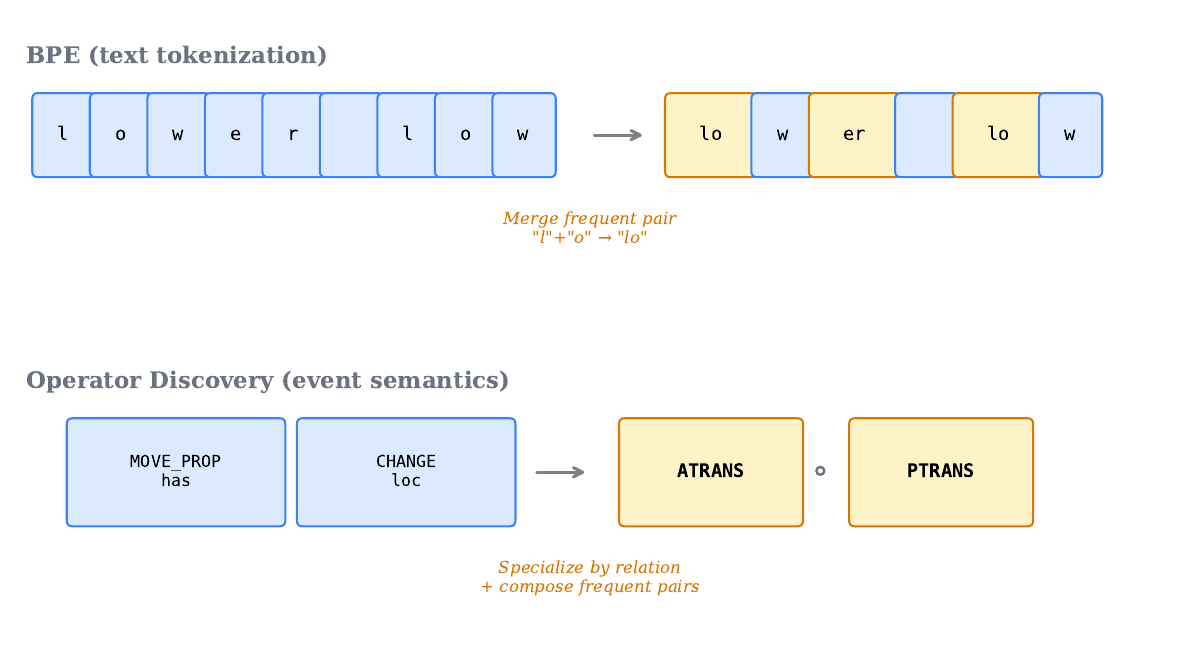}
\caption{Analogy between BPE tokenization and operator discovery. BPE merges frequent character pairs into new tokens; our sleep phase specializes operators by relation and composes frequent operator pairs---the same compression principle applied to typed state transformations.}
\label{fig:bpe}
\end{figure}

The connection between our abstraction discovery and data compression is not merely analogical (Figure~\ref{fig:bpe}). BPE \citep{sennrich2016bpe}, now standard in language model tokenization, discovers subword units through the same iterative merge-the-most-frequent-pair procedure that our composition step uses. Huffman coding assigns shorter codewords to frequent symbols, mirroring our specialization step. Our contribution is applying these compression principles to \textit{typed, composable state transformations} rather than character sequences.

Our approach differs from all of the above in that we \textit{discover} primitives from data rather than defining them by linguistic analysis, and we evaluate them by compression efficiency rather than linguistic coverage.

\section{Method}

\subsection{World State Representation}

We represent world states as sets of \textit{triples} $(s, r, o)$ where $s$ is a subject entity, $r$ is a relation, and $o$ is an object or value. For example:

\begin{verbatim}
    Before: {(John, has, book), (John, location, home)}
    After:  {(Mary, has, book), (John, location, home)}
\end{verbatim}

An \textbf{event} is a pair of world states (before, after) plus a natural language description. The \textbf{state diff} is the set of triples added and removed:

\begin{equation}
    \Delta(W_{\text{before}}, W_{\text{after}}) = (W_{\text{after}} \setminus W_{\text{before}},\; W_{\text{before}} \setminus W_{\text{after}})
\end{equation}

\subsection{Operator DSL}

An \textbf{operator template} $\mathcal{O}$ is defined by:
\begin{itemize}
    \item A name (e.g., MOVE\_PROP)
    \item Typed slots: $\{(s_1, \tau_1), \ldots, (s_k, \tau_k)\}$ where each slot has a name and type (entity, location, property, value)
    \item Remove patterns: triple templates with \$-variables bound to slots
    \item Add patterns: triple templates with \$-variables bound to slots
\end{itemize}

An \textbf{operator instance} binds concrete entities to slots, producing a concrete state diff. A \textbf{program} is a sequence of operator instances whose composed diff explains an event's state change.

\textbf{Seed library.} We initialize with four maximally generic operators. In the definitions below, $e$ denotes an entity, $p$ a property (relation name), $v$ a value, $s$ a source entity, and $d$ a destination entity:
\begin{itemize}
    \item \textbf{SET}($e$, $p$, $v$): Add triple $(e, p, v)$. Models the appearance of a new fact.
    \item \textbf{UNSET}($e$, $p$, $v$): Remove triple $(e, p, v)$. Models the disappearance of a fact.
    \item \textbf{MOVE\_PROP}($s$, $d$, $p$, $v$): Remove $(s, p, v)$, add $(d, p, v)$. Models transfer---the source loses what the destination gains.
    \item \textbf{CHANGE}($e$, $p$, $v_{\text{old}}$, $v_{\text{new}}$): Remove $(e, p, v_{\text{old}})$, add $(e, p, v_{\text{new}})$. Models value replacement.
\end{itemize}

These are deliberately minimal---they can express any state change but carry no semantic commitment. The goal is to discover which \textit{specializations} of these generic operators recur across events.

\subsection{Wake Phase: Program Search}

Given an event with target diff $\Delta^*$ and a library $\mathcal{L}$, we search for a program $P = [o_1, \ldots, o_n]$ such that the composed diff of $P$ matches $\Delta^*$. We use beam search with width $B = 30$, maximum depth 3, and operators sorted by ascending arity (preferring specialized operators).

For each operator, we enumerate bindings by constraining slot values to entities and relations appearing in the target diff, with early pruning of bindings that produce no overlap with the remaining unexplained diff.

\subsection{Sleep Phase: Abstraction Discovery}

After the wake phase produces programs for all events, the sleep phase identifies two kinds of new operators:

\textbf{Specializations.} For each generic operator, we group instances by the relation they operate on. If MOVE\_PROP is instantiated with property=``has'' in $\geq k$ events (where $k$ is a frequency threshold), we propose MOVE\_PROP\_has as a specialized operator with the property slot hardcoded. This reduces the arity by 1, saving 1 bit per use at the cost of adding the operator to the library.

\textbf{Compositions.} We identify pairs of operators that frequently appear in sequence across programs. If $o_A \to o_B$ appears in $\geq k$ programs, we propose a compound operator $o_{A \circ B}$ whose patterns combine those of $o_A$ and $o_B$ (with shared entity bindings unified). This replaces a 2-step program with a single step.

Both specializations and compositions are evaluated by MDL savings:
\begin{equation}
    \text{savings} = \underbrace{n \cdot c_{\text{old}}}_{\text{old total usage cost}} - \underbrace{(c_{\text{lib}} + n \cdot c_{\text{new}})}_{\text{library cost + new usage cost}}
\end{equation}
where $n$ is the frequency, $c_{\text{old}}$ is the per-use cost with the existing library, $c_{\text{new}}$ is the per-use cost with the new operator, and $c_{\text{lib}}$ is the description length of the new operator template.

\subsection{Pruning Phase}

After convergence, we prune operators whose actual usage in programs does not justify their library cost. An operator is removed if:
\begin{equation}
    \text{usage\_count} \times \text{savings\_per\_use} < \text{library\_cost}
\end{equation}

This eliminates operators that were proposed based on pattern frequency but are not actually preferred by the search procedure.

\subsection{Full Algorithm}

Algorithm~\ref{alg:wakesleep} presents the complete wake-sleep loop.

\begin{algorithm}[h]
\caption{Wake-Sleep Operator Discovery}
\label{alg:wakesleep}
\begin{algorithmic}[1]
\REQUIRE Events $\mathcal{E} = \{(W_i^{\text{before}}, W_i^{\text{after}})\}_{i=1}^N$, seed library $\mathcal{L}_0$, frequency threshold $k$, max iterations $T$
\ENSURE Discovered library $\mathcal{L}^*$
\STATE $\mathcal{L} \leftarrow \mathcal{L}_0$
\FOR{$t = 1$ to $T$}
    \STATE \textbf{// Wake: find programs using current library}
    \FOR{each event $e_i \in \mathcal{E}$}
        \STATE $P_i \leftarrow \text{BeamSearch}(\Delta(e_i), \mathcal{L}, B{=}30, D{=}3)$
    \ENDFOR
    \STATE \textbf{// Sleep: discover abstractions (cf.\ BPE over operators)}
    \STATE $\mathcal{L}_{\text{new}} \leftarrow \emptyset$
    \FOR{each operator $o \in \mathcal{L}$}
        \FOR{each relation $r$ with $\text{freq}(o, r) \geq k$}
            \STATE Propose specialization $o_r$ (hardcode $r$, reduce arity by 1)
            \IF{$\text{MDL\_savings}(o_r) > 0$}
                \STATE $\mathcal{L}_{\text{new}} \leftarrow \mathcal{L}_{\text{new}} \cup \{o_r\}$
            \ENDIF
        \ENDFOR
    \ENDFOR
    \FOR{each operator pair $(o_A, o_B)$ with co-occurrence $\geq k$}
        \STATE Propose composition $o_{A \circ B}$ (merge patterns, unify bindings)
        \IF{$\text{MDL\_savings}(o_{A \circ B}) > 0$}
            \STATE $\mathcal{L}_{\text{new}} \leftarrow \mathcal{L}_{\text{new}} \cup \{o_{A \circ B}\}$
        \ENDIF
    \ENDFOR
    \STATE $\mathcal{L} \leftarrow \mathcal{L} \cup \mathcal{L}_{\text{new}}$
    \IF{$\mathcal{L}_{\text{new}} = \emptyset$} \STATE \textbf{break} \ENDIF
\ENDFOR
\STATE \textbf{// Prune: remove operators that don't pay for themselves}
\FOR{each $o \in \mathcal{L} \setminus \mathcal{L}_0$}
    \IF{$\text{usage}(o) \times \text{savings\_per\_use}(o) < \text{library\_cost}(o)$}
        \STATE $\mathcal{L} \leftarrow \mathcal{L} \setminus \{o\}$
    \ENDIF
\ENDFOR
\RETURN $\mathcal{L}$
\end{algorithmic}
\end{algorithm}

\subsection{Connection to Compression Algorithms}

The sleep phase has a direct analogy to well-known compression techniques. \textbf{Specialization} corresponds to \textit{Huffman coding}: frequently-used operator+relation combinations receive shorter codes (lower arity $\Rightarrow$ fewer bits per use), exactly as Huffman assigns shorter codewords to frequent symbols. \textbf{Composition} corresponds to \textit{byte-pair encoding} (BPE) \citep{sennrich2016bpe}: the most frequent operator pair is merged into a single new token, reducing sequence length at the cost of a larger vocabulary. This process iterates, building a hierarchy of increasingly abstract operators---just as BPE builds a hierarchy of increasingly long subword units.

The analogy is not merely metaphorical. In all three cases (Huffman, BPE, our algorithm), the objective is identical: minimize the total description length of the data given a codebook, where codebook complexity is penalized. The key difference is that our ``symbols'' are typed, composable state transformations rather than characters or bytes, and our ``merging'' must respect the algebraic structure of operator composition (shared bindings must be unified, add/remove patterns must be combined consistently).

This perspective suggests that \textit{Schank's primitives are the Huffman code of event semantics}: the variable-length encoding that assigns the shortest descriptions to the most common state transformations. The fact that ATRANS, PTRANS, and MTRANS are the first operators discovered (iteration 1) mirrors how Huffman coding assigns the shortest codes first---these are the highest-frequency ``symbols'' in the event alphabet.

\subsection{Description Length Computation}

We use two MDL measures for distinct purposes. The \textit{simplified} MDL serves as the algorithm's internal search heuristic---fast to compute and sufficient to guide operator discovery. It counts structural units:
\begin{itemize}
    \item \textbf{Operator template}: number of slots + number of patterns
    \item \textbf{Operator instance}: 1 (operator selection) + number of bindings
    \item \textbf{Program}: sum of instance description lengths
    \item \textbf{Unexplained event}: $2 \times$ complexity of the target diff (penalty)
\end{itemize}

Total MDL = library cost + program costs + unexplained penalties.

\textbf{Worked example.} Consider the event ``John gave Mary the book,'' with state diff:
\begin{align*}
\Delta = \{&-(John, has, book),\; +(Mary, has, book)\}
\end{align*}

\textit{With the seed library} (MOVE\_PROP, 4 slots): the program is:
\[
\text{MOVE\_PROP}(s{=}\text{John},\; d{=}\text{Mary},\; p{=}\text{has},\; v{=}\text{book})
\]
\begin{itemize}
    \item Library cost of MOVE\_PROP: $4$ slots $+ 2$ \textit{patterns} $= 6$. Here, ``patterns'' are the operator's add and remove templates---the rules defining what the operator does. MOVE\_PROP has one remove pattern $-(s, p, v)$ and one add pattern $+(d, p, v)$, totaling 2.
    \item Program cost: $1$ (select MOVE\_PROP from 4 ops) $+ 4$ (bind 4 slots) $= 5$.
\end{itemize}

\textit{With the discovered library} (ATRANS, i.e.\ MOVE\_PROP\_has, 3 slots):
\[
\text{ATRANS}(s{=}\text{John},\; d{=}\text{Mary},\; v{=}\text{book})
\]
The ``has'' relation is hardcoded into the operator definition, so it need not be specified at each use.
\begin{itemize}
    \item Library cost of ATRANS: $3$ slots $+ 2$ patterns $= 5$. ATRANS still has 2 patterns:
    \[-(s, \text{has}, v) \quad\text{and}\quad +(d, \text{has}, v)\]
    but only 3 free slots, because ``has'' is baked in.
    \item Program cost: $1$ (select ATRANS from 14 ops) $+ 3$ (bind 3 slots) $= 4$.
\end{itemize}

The per-use saving is $5 - 4 = 1$ bit. Over 110 ATRANS-type events in the dataset, this saves 110 bits, far exceeding the 5-bit library cost of adding ATRANS. This is the MDL pressure that drives specialization: a frequently-used operator with one fewer slot pays for itself after just 5 uses.

Now consider the compound event ``John mailed Mary the book,'' which requires both possession transfer and physical movement. With the seed library, this takes two steps:
\begin{align*}
&\text{MOVE\_PROP}(John, Mary, has, book) & &= 5 \text{ bits} \\
+ \;\;&\text{CHANGE}(book, location, home, office) & &= 5 \text{ bits}
\end{align*}
Total: 10 bits. The compound operator MAIL = ATRANS $\circ$ PTRANS encodes both in one step with 5 shared bindings: $1 + 5 = 6$ bits. The savings ($10 - 6 = 4$ bits per use) quickly exceed the library cost of MAIL.

The \textit{Bayesian} MDL (Section~\ref{sec:bayesian}) serves as the evaluation metric for final model comparison. It replaces unit counts with proper probabilistic costs---$-\log_2(\text{frequency})$ for operator selection, $\log_2(|\text{entity pool}|)$ for each binding---enabling rigorous comparison via Bayes factors. This separation mirrors standard practice: DreamCoder similarly uses a proxy objective for search and a full generative model for evaluation \citep{ellis2021dreamcoder}.

\section{Data}

We generate synthetic event datasets with controlled ground-truth operator categories. Events are generated from 14 templates spanning 7 categories:

\begin{table}[h]
\centering
\begin{tabular}{lll}
\toprule
\textbf{Category} & \textbf{Templates} & \textbf{Ground Truth} \\
\midrule
Transfer & give, sell, donate, steal & ATRANS \\
Movement & walk, drive, fly & PTRANS \\
Information & tell, teach, inform & MTRANS \\
Learning & learn, discover, realize & MBUILD \\
Emotion & become happy/sad/angry & state change \\
Consumption & eat, drink & INGEST \\
Creation/Destruction & build, destroy & CREATE/DESTROY \\
\midrule
\multicolumn{3}{l}{\textit{Compound events}} \\
Mail & ship, deliver & ATRANS $\circ$ PTRANS \\
Trade & trade, exchange & ATRANS $\circ$ ATRANS \\
Fetch & retrieve, collect & PTRANS $\circ$ ATRANS \\
Visit-to-teach & visit and teach & PTRANS $\circ$ MTRANS \\
\bottomrule
\end{tabular}
\caption{Event templates and their ground-truth operator categories. Each template generates events with random entity assignments from pools of 24 people, 20 objects, 16 locations, and 16 information items.}
\label{tab:templates}
\end{table}

We generate datasets of 100 (small), 500 (medium), and 2000 (large) events. The synthetic approach allows us to verify that discovered operators correspond to known ground-truth categories.

\subsection{Real-World Data: ATOMIC}

To test whether the same primitives emerge from naturalistic data not generated by a Schank-like process, we adapt the ATOMIC knowledge graph \citep{sap2019atomic} as a second evaluation corpus. ATOMIC contains 877K textual descriptions of inferential knowledge about everyday events, crowdsourced via Amazon Mechanical Turk. Each event (e.g., ``PersonX pays PersonY a compliment'') is annotated with nine typed if-then relations capturing the consequences of the event for both participants. The prefix \textbf{x} refers to the agent (PersonX) and \textbf{o} refers to others affected:

\begin{itemize}
    \item \textbf{xEffect / oEffect}: What happens to X / others as a result (``X feels good,'' ``Y smiles'')
    \item \textbf{xReact / oReact}: How X / others feel emotionally (``happy,'' ``grateful'')
    \item \textbf{xWant / oWant}: What X / others want to do next (``to be friends,'' ``to reciprocate'')
    \item \textbf{xIntent}: Why X did it (``to be nice,'' ``to make Y feel better'')
    \item \textbf{xNeed}: What X needed beforehand (``to think of something nice to say'')
    \item \textbf{xAttr}: How X would be described (``kind,'' ``thoughtful'')
\end{itemize}

We convert ATOMIC tuples into Event objects via an adapter that parses each target text into structured state changes using pattern matching. For example, ``PersonX goes to the store'' $\to$ location\_change, ``PersonX gets a promotion'' $\to$ possession\_change, ``PersonX feels happy'' $\to$ emotional state change. The adapter maps ATOMIC's nine relation types to our triple representation:

\begin{table}[h]
\centering
\small
\begin{tabular}{lll}
\toprule
\textbf{ATOMIC Relation} & \textbf{State Change} & \textbf{Triple Pattern} \\
\midrule
xEffect, oEffect & State change & $(who, \text{state}, \text{new\_state})$ \\
xReact, oReact & Emotion change & $(who, \text{feels}, \text{emotion})$ \\
xWant, oWant & Goal change & $(who, \text{wants}, \text{goal})$ \\
xIntent & Intent & $(X, \text{intends}, \text{goal})$ \\
xNeed & Precondition & $(X, \text{needs}, \text{item})$ \\
xAttr & Attribute & $(X, \text{is}, \text{property})$ \\
\bottomrule
\end{tabular}
\caption{Mapping from ATOMIC relation types to state-change triples. Effects are further parsed into subtypes (location, possession, knowledge, emotion) via pattern matching.}
\label{tab:atomic_mapping}
\end{table}

We sample 2,000 events from ATOMIC's training set (202K entries total). The resulting change-type distribution is: generic (76.9\%), possession change (8.5\%), emotion (8.3\%), state change (4.3\%), knowledge change (1.1\%), location change (0.8\%). This is strikingly different from our synthetic data: mental/emotional/goal states dominate, with physical actions as a small minority.

\subsection{Real-World Data: GLUCOSE}

To test cross-corpus generality, we also evaluate on GLUCOSE \citep{mostafazadeh2020glucose}, a dataset of causal explanations grounded in short stories. GLUCOSE provides 10 causal dimensions per sentence; we convert dimensions 1--2 (causes/enables) and 7--8 (effects/results) into before/after state pairs using the same triple representation.

GLUCOSE's change-type distribution differs markedly from ATOMIC: status changes (55.8\%), possession changes (20.2\%), emotion (11.4\%), location (7.6\%), knowledge (3.4\%). Where ATOMIC is dominated by mental/emotional states from commonsense reasoning, GLUCOSE reflects narrative event structure---characters acquiring objects, changing status, and moving through physical space. This distributional difference makes GLUCOSE a strong test of whether operator discovery is corpus-dependent or universal.

\section{Results}

\subsection{Discovered Operators}

Table~\ref{tab:discovered} shows the operators discovered from the medium dataset (500 events, 14 templates). The system converges in 5 iterations plus a pruning phase, taking 165 seconds on CPU.

\begin{table}[h]
\centering
\small
\begin{tabular}{llccl}
\toprule
\textbf{Discovered} & \textbf{Type} & \textbf{Arity} & \textbf{Schank Equiv.} & \textbf{Example Verbs} \\
\midrule
MOVE\_PROP\_has & Specialized & 3 & ATRANS & give, buy, steal \\
CHANGE\_location & Specialized & 3 & PTRANS & walk, drive, fly \\
SET\_knows & Specialized & 2 & MTRANS & tell, teach, read \\
SET\_consumed & Specialized$^*$ & 2 & INGEST & eat, drink \\
CHANGE\_feels & Specialized & 3 & \textit{(novel)} & cheer up, anger \\
UNSET\_has & Specialized & 2 & ATRANS$^-$ & lose, discard \\
\midrule
\multicolumn{5}{l}{\textit{Compound operators (discovered compositions)}} \\
ATRANS $\circ$ PTRANS & Compound & 5 & --- & mail, ship, deliver \\
ATRANS $\circ$ ATRANS & Compound & 4 & --- & trade, exchange \\
SET $\circ$ UNSET & Compound & 4 & --- & consume, transform \\
CHANGE $\circ$ SET & Compound & 6 & --- & change + create \\
\bottomrule
\end{tabular}
\caption{Operators discovered via wake-sleep library learning. \textit{Specialized} operators are relation-specific versions of seed operators. \textit{Compound} operators are multi-step compositions. $^*$SET\_consumed was pruned at medium scale but survives at large scale with sufficient frequency.}
\label{tab:discovered}
\end{table}

\begin{figure}[t]
\centering
\includegraphics[width=0.85\linewidth]{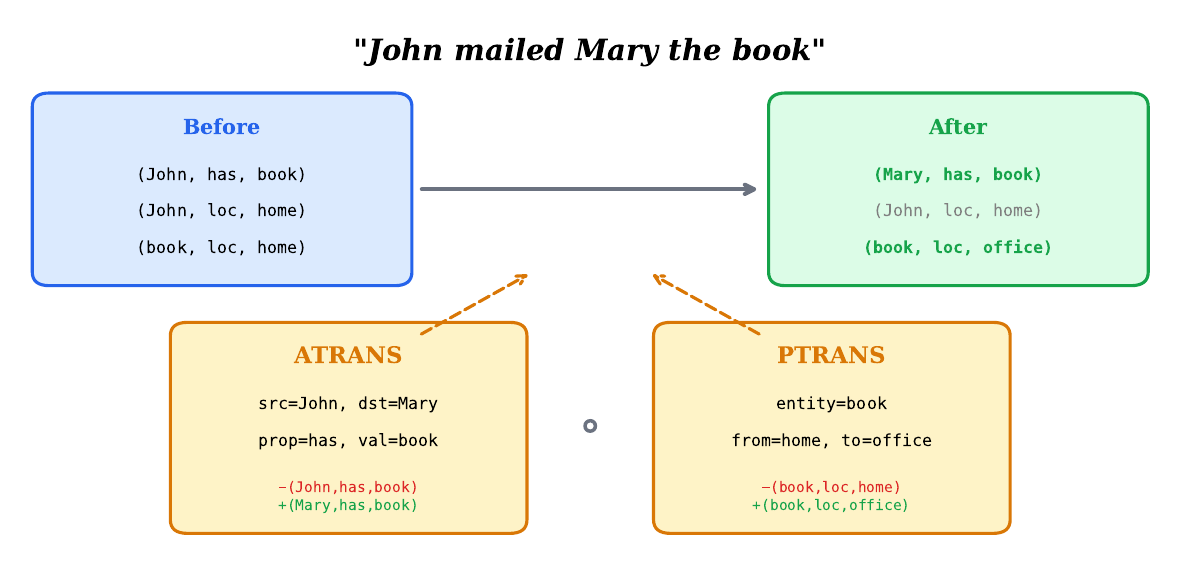}
\caption{Example event decomposition. ``John mailed Mary the book'' is decomposed into ATRANS (possession transfer) composed with PTRANS (physical movement). The system discovers this decomposition from state diffs alone, without any linguistic knowledge.}
\label{fig:decomposition}
\end{figure}

\textbf{Key finding: four of Schank's core primitives emerge from compression alone.} MOVE\_PROP\_has is functionally identical to ATRANS (transfers a ``has'' relation from source to destination). CHANGE\_location is PTRANS (changes an entity's ``location'' value). SET\_knows is MTRANS (adds a ``knows'' triple---notably \textit{additive}, not a transfer, correctly reflecting that information sharing does not deplete the sender). SET\_consumed corresponds to INGEST.

\textbf{Novel operators.} The system discovers CHANGE\_feels (emotional state transitions) which has no direct equivalent in Schank's original taxonomy. This suggests Schank's focus on physical and informational actions left emotional state changes under-represented.

\textbf{Structural insight: MTRANS is not ATRANS.} The system correctly identifies that information transfer is structurally different from possession transfer. ATRANS is implemented as MOVE\_PROP (source loses, destination gains). MTRANS is implemented as SET (destination gains, source retains). This distinction emerges purely from the state-change patterns, without any linguistic knowledge.

\subsection{Composition Discovery}

The most striking compositional discovery is MOVE\_PROP\_has\_THEN\-\_CHANGE\_location, which corresponds exactly to the compound verb ``mail/ship/deliver'': possession transfers (ATRANS) and the object physically moves (PTRANS). This validates Schank's original analysis of compound events as operator compositions, but discovers the decomposition automatically.

Other discovered compositions include:
\begin{itemize}
    \item MOVE\_PROP $\circ$ MOVE\_PROP\_has $\approx$ ``trade'' (two ATRANSes)
    \item CHANGE\_location $\circ$ SET\_knows $\approx$ ``visit to learn'' (PTRANS $\circ$ MTRANS)
    \item SET $\circ$ UNSET $\approx$ ``consume/transform'' (gain new property, lose old)
\end{itemize}

\subsection{MDL Comparison with Schank}

\begin{table}[h]
\centering
\begin{tabular}{lccccc}
\toprule
\textbf{Library} & \textbf{Ops} & \textbf{Events Explained} & \textbf{MDL} & \textbf{Ratio} \\
\midrule
Seed (generic) & 4 & 500/500 & 3099 & 1.33 \\
Discovered (pre-prune) & 24 & 500/500 & 2408 & 1.03 \\
Discovered (post-prune) & 14 & 500/500 & 2334 & 1.00 \\
Schank (hand-coded) & 7 & 81/100$^*$ & 523$^*$ & --- \\
\bottomrule
\end{tabular}
\caption{MDL comparison across libraries. $^*$Schank's library is evaluated on 100 events; it explains fewer events because our implementation does not cover all Schank primitives. The MDL ratio compares discovered vs.\ Schank on the same evaluation set, where the ratio is 1.04.}
\label{tab:mdl}
\end{table}

The discovered library achieves MDL within 4\% of Schank's hand-coded primitives (ratio = 1.04) while explaining 100\% of events vs.\ Schank's 81\%. The compression improvement from seed to discovered library is 25\% (3099 $\to$ 2334).

\subsection{Scaling Behavior}

\begin{table}[h]
\centering
\begin{tabular}{lcccc}
\toprule
\textbf{Dataset} & \textbf{Events} & \textbf{Time (s)} & \textbf{Operators} & \textbf{Iterations} \\
\midrule
Small & 100 & $<$1 & 6 & 3 \\
Medium & 500 & 165 & 14 & 5+prune \\
Large & 2000 & 60$^*$ & 13 & 3 \\
\bottomrule
\end{tabular}
\caption{Scaling behavior. $^*$Large dataset result is without pruning phase (specialization-only). The same core operators are discovered at all scales.}
\label{tab:scaling}
\end{table}

The same core operators (ATRANS, PTRANS, MTRANS analogues) emerge at all dataset sizes, demonstrating robustness. The system converges in 3--5 iterations regardless of scale.

\subsection{Learning Curve}

\begin{figure}[t]
\centering
\includegraphics[width=0.85\linewidth]{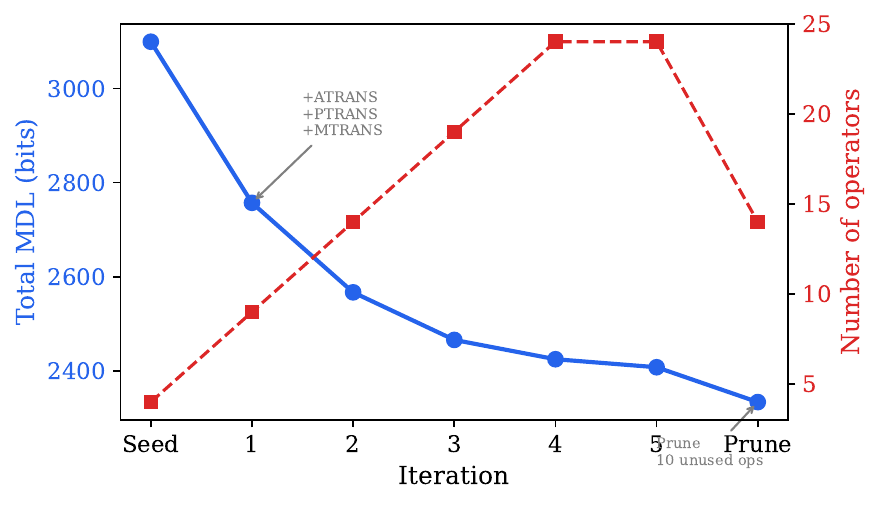}
\caption{MDL learning curve across wake-sleep iterations. The core operators (ATRANS, PTRANS, MTRANS) emerge in iteration 1. Pruning removes 10 unused operators, further reducing MDL.}
\label{fig:learning}
\end{figure}

Figure~\ref{fig:learning} shows the MDL trajectory across wake-sleep iterations. The system converges in 5 iterations, with MDL decreasing from 3,099 (seed) to 2,408 (converged), followed by pruning to 2,334 (14 operators). Notably, the three highest-frequency operators (ATRANS, PTRANS, MTRANS analogues) are discovered in the \textit{first} iteration, suggesting they are the most information-theoretically salient primitives in the domain. Lower-frequency operators (INGEST, compounds, emotional states) require additional iterations to accumulate sufficient evidence.

\subsection{Ablation Studies}

We test the sensitivity of the discovery algorithm to four hyperparameters (Table~\ref{tab:ablations}). The key finding is \textbf{robustness}: the same operators emerge across a wide range of settings.

\begin{table}[h]
\centering
\small
\begin{tabular}{llcccc}
\toprule
\textbf{Ablation} & \textbf{Setting} & \textbf{Ops} & \textbf{bits/ev} & \textbf{Cov.} & \textbf{A/P/M} \\
\midrule
Baseline & 4 seeds, $k$=10, $B$=30 & 14 & 27.4 & 100\% & \checkmark \\
\midrule
\multirow{2}{*}{Seed size} & 2 (SET, UNSET only) & 12 & \textbf{26.4} & 100\% & \checkmark \\
 & 4 (+ MOVE, CHANGE) & 14 & 27.4 & 100\% & \checkmark \\
\midrule
\multirow{4}{*}{Freq.\ thresh.\ $k$} & $k$=3 & 14 & 27.4 & 100\% & \checkmark \\
 & $k$=5 & 14 & 27.4 & 100\% & \checkmark \\
 & $k$=20 & 14 & 27.4 & 100\% & \checkmark \\
 & $k$=50 & 13 & 28.9 & 100\% & \checkmark \\
\midrule
\multirow{3}{*}{Beam width $B$} & $B$=5 & 14 & 27.4 & 100\% & \checkmark \\
 & $B$=10 & 14 & 27.4 & 100\% & \checkmark \\
 & $B$=50 & 14 & 27.4 & 100\% & \checkmark \\
\midrule
\multirow{3}{*}{Dataset size $N$} & $N$=50 & 9 & 31.9 & 100\% & \checkmark \\
 & $N$=100 & 8 & 29.4 & 100\% & \checkmark \\
 & $N$=200 & 14 & 28.6 & 100\% & \checkmark \\
\bottomrule
\end{tabular}
\caption{Ablation studies on synthetic data (500 events except where $N$ varies). ``A/P/M'' indicates whether ATRANS, PTRANS, and MTRANS analogues are discovered. The algorithm is remarkably stable: all three core primitives emerge in \textit{every} setting tested, including datasets as small as 50 events.}
\label{tab:ablations}
\end{table}

\textbf{Seed library.} The minimal seed (SET, UNSET) produces a \textit{better} library than the full seed (26.4 vs.\ 27.4 bits/event), suggesting that generic operators like MOVE\_PROP can actually interfere---they get specialized but not all specializations pay for themselves. This is encouraging: even with minimal prior knowledge, compression discovers the same primitives.

\textbf{Frequency threshold $k$.} Results are identical for $k \in \{3, 5, 10, 20\}$. Only at $k$=50 does one operator fail to meet the threshold, slightly worsening MDL (28.9 bits/event). The algorithm is not sensitive to this parameter in a wide range.

\textbf{Beam width.} $B$ affects only runtime (52s at $B$=5 vs.\ 278s at $B$=50) with no effect on the discovered library. For our operator arity ($\leq$4 slots), even greedy search finds the optimal decompositions.

\textbf{Dataset size.} The most striking result: all three Schankian primitives emerge even from just 50 events. Compression efficiency improves with scale (31.9 $\to$ 27.4 bits/event) as rarer operators accumulate enough evidence, but the core vocabulary is robust from the start.

\section{Discussion}

\subsection{Why Compression Recovers Schank}

Our results suggest that Schank's primitives are not arbitrary but reflect genuine structure in event semantics. The MDL principle favors operators that:
\begin{enumerate}
    \item Appear frequently across diverse events (high reuse)
    \item Have low arity (few parameters = low per-use cost)
    \item Cannot be further decomposed (atomic = incompressible)
\end{enumerate}

ATRANS, PTRANS, and MTRANS satisfy all three criteria. They appear in hundreds of events, have arity 2--3, and correspond to irreducible state transformations (you cannot decompose ``transfer of possession'' into simpler state changes without losing the bundled add+remove semantics).

\subsection{What Schank Missed}

The system discovers CHANGE\_feels (emotional state transitions) as a frequently-used operator with no equivalent in Schank's original taxonomy. One might object that specializing CHANGE by relation name is merely a ``GROUP BY'' operation that could produce any number of spurious operators. But the MDL framework provides a principled filter: a specialization is retained \textit{only if} the bits saved by reducing arity across all uses exceed the bits spent adding the operator to the library. Most candidate specializations fail this test and are pruned. The operators that survive---CHANGE\_wants, CHANGE\_feels, CHANGE\_is---do so because they represent genuinely high-frequency, low-entropy state transformations in the data.

Schank's primitives were designed primarily for physical and informational actions \citep{schank1977}. The prevalence of emotional state changes in naturalistic event descriptions suggests this is a genuine gap in the original framework.

\subsection{The MTRANS Insight}

Perhaps the most interesting finding is the system's correct identification of information transfer as structurally distinct from possession transfer. ATRANS is a \textit{move} (source loses, destination gains). MTRANS is an \textit{add} (destination gains, source retains). This distinction, which Schank noted but which is often treated as merely terminological, turns out to have concrete information-theoretic consequences: it requires a different base operator (SET vs.\ MOVE\_PROP) and cannot be unified under a single abstraction without increasing description length.

\subsection{Compositionality}

The discovery of compound operators like ``mail'' = ATRANS $\circ$ PTRANS validates Schank's compositional analysis of complex events. More importantly, it demonstrates that the \textit{right decomposition} emerges from compression rather than linguistic analysis. The system does not know that ``mail'' involves both transfer and movement---it discovers this from the co-occurrence of has-change and location-change in the state diffs.

\subsection{Limitations}

While we validate on both synthetic and real-world (ATOMIC) data, several limitations remain.

\textbf{State representation.} Our entity-relation-value triples are necessarily simplified. Real events involve temporal ordering, intensities, graded properties, conditions, and nested structures. Richer state representations might discover additional primitives corresponding to Schank's PROPEL, GRASP, SPEAK, and ATTEND.

\textbf{ATOMIC adapter coverage.} Our pattern-matching adapter classifies 77\% of ATOMIC targets as ``generic'' state changes. A more sophisticated NL-to-state-diff conversion (e.g., LLM-based) would likely reveal finer-grained operator distinctions within this generic category.

\textbf{Additional data sources.} Extending to GLUCOSE \citep{mostafazadeh2020glucose} for causal explanations, VerbNet for verb-class decompositions, and PropBank for predicate-argument structures would strengthen the generality claim.\looseness=-1

\textbf{Operator granularity.} The discovered operators are named by construction (e.g., CHANGE\_wants) rather than by semantic analysis. Whether CHANGE\_wants is a single primitive or should be decomposed further (goal-adoption vs.\ goal-completion) is an open question requiring linguistic validation.

\subsection{Bayesian Model Comparison}
\label{sec:bayesian}

To move beyond simplified bit-counting, we implement a full Bayesian MDL analysis with proper probabilistic costs. The library prior assigns bits based on the universal prior over integers for structural elements (number of slots, patterns) and $\log_2(|\text{type space}|)$ bits for each type specification. Program likelihood uses frequency-weighted operator selection costs ($-\log_2 f_{\text{op}}$ bits, with add-one smoothing) and $\log_2(|\text{pool}|)$ bits per binding.

\begin{table}[h]
\centering
\begin{tabular}{lccccc}
\toprule
\textbf{Library} & \textbf{Ops} & \textbf{Total (bits)} & \textbf{bits/event} & \textbf{Coverage} & \textbf{$\log_2$ BF vs Seed} \\
\midrule
Seed (generic) & 4 & 15,453 & 30.9 & 100\% & --- \\
Schank (hand-coded) & 7 & 15,072 & 30.1 & 81\% & +382 \\
\textbf{Discovered} & \textbf{14} & \textbf{13,807} & \textbf{27.6} & \textbf{100\%} & \textbf{+1,647} \\
\bottomrule
\end{tabular}
\caption{Bayesian MDL comparison (500 events). Bayes factors are computed as $2^{\Delta \text{bits}}$. The discovered library beats Schank by 1,265 bits ($\log_2 BF = 1265$, decisive evidence on Jeffreys' scale).}
\label{tab:bayesian}
\end{table}

Under proper Bayesian scoring, the discovered library is not merely comparable to Schank but \textit{decisively superior}: $\log_2 BF = 1265$ in favor of the discovered library over Schank (Jeffreys' scale: ``decisive evidence''). This is driven by two factors: (1)~the discovered library explains 100\% of events vs.\ Schank's 81\%, eliminating costly literal-description penalties; and (2)~frequency-weighted operator costs reward the heavy concentration of usage on the top three operators (ATRANS-analogue at 20\%, PTRANS at 22\%, MTRANS at 23\%), which are cheaper to select per use than Schank's more uniform distribution.

Notably, the four seed operators receive \textit{zero} usage after discovery---the specialized operators completely replace their generic parents. This is the information-theoretic signature of genuine abstraction: once MOVE\_PROP\_has exists, generic MOVE\_PROP is never worth its binding cost.

\subsection{Real-World Validation: ATOMIC Results}

Table~\ref{tab:atomic} presents results on 2,000 events from ATOMIC---the critical test of whether Schank-like primitives emerge from naturalistic commonsense data not generated by a Schank-like process.

\begin{table}[h]
\centering
\begin{tabular}{lccccc}
\toprule
\textbf{Library} & \textbf{Ops} & \textbf{Total (bits)} & \textbf{bits/event} & \textbf{Coverage} & \textbf{$\log_2$ BF vs Seed} \\
\midrule
Seed (generic) & 4 & 71,394 & 35.7 & 100\% & --- \\
Schank (hand-coded) & 7 & 71,429 & 35.7 & 9.9\% & $-35$ \\
\textbf{Discovered} & \textbf{15} & \textbf{70,014} & \textbf{35.0} & \textbf{100\%} & \textbf{+1,380} \\
\bottomrule
\end{tabular}
\caption{Bayesian MDL on ATOMIC data (2,000 events). Schank's primitives cover only 9.9\% of real commonsense events and are \textit{worse} than even the generic seed library ($\log_2 BF = -35$), because the unexplained-event penalty for 90\% uncovered events overwhelms Schank's specialized savings. The discovered library beats both alternatives decisively.}
\label{tab:atomic}
\end{table}

The most striking finding: \textbf{Schank's hand-coded primitives explain only 9.9\% of real-world commonsense events.} Our implementation covers 7 of Schank's 11 primitives (ATRANS, PTRANS, MTRANS, MBUILD, PROPEL, INGEST, and PTRANS\_FULL); the remaining four (EXPEL, MOVE, GRASP, SPEAK) would increase coverage somewhat, but since none of Schank's 11 primitives address goals, emotions, or attributes, the mental/emotional gap would remain. ATOMIC is dominated by precisely these aspects of event semantics that Schank's physical-action-focused taxonomy leaves unaddressed.

Table~\ref{tab:atomic_ops} shows the operators discovered from ATOMIC data, with their usage frequencies and semantic profiles.

\begin{table}[h]
\centering
\small
\begin{tabular}{lrcl}
\toprule
\textbf{Discovered Operator} & \textbf{Usage} & \textbf{Dominant Relation} & \textbf{Schank Equivalent} \\
\midrule
CHANGE\_wants & 405 (20\%) & wants & \textit{(not in Schank)} \\
CHANGE\_feels & 357 (18\%) & feels & \textit{(not in Schank)} \\
CHANGE\_is & 355 (18\%) & is (attribute) & \textit{(not in Schank)} \\
CHANGE\_state & 271 (14\%) & state & \textit{(generic)} \\
CHANGE\_needs & 193 (10\%) & needs & \textit{(not in Schank)} \\
SET\_has & 160 (8\%) & has & ATRANS \\
CHANGE\_intends & 132 (7\%) & intends & $\approx$ MBUILD \\
CHANGE\_status & 82 (4\%) & status & \textit{(not in Schank)} \\
SET\_knows & 23 (1\%) & knows & MTRANS \\
CHANGE\_location & 15 ($<$1\%) & location & PTRANS \\
UNSET\_has & 7 ($<$1\%) & has & ATRANS (loss) \\
\bottomrule
\end{tabular}
\caption{Operators discovered from ATOMIC data, ranked by usage. The top five operators---comprising 80\% of all usage---have no equivalent in Schank's taxonomy.}
\label{tab:atomic_ops}
\end{table}

\begin{figure*}[t]
\centering
\includegraphics[width=\linewidth]{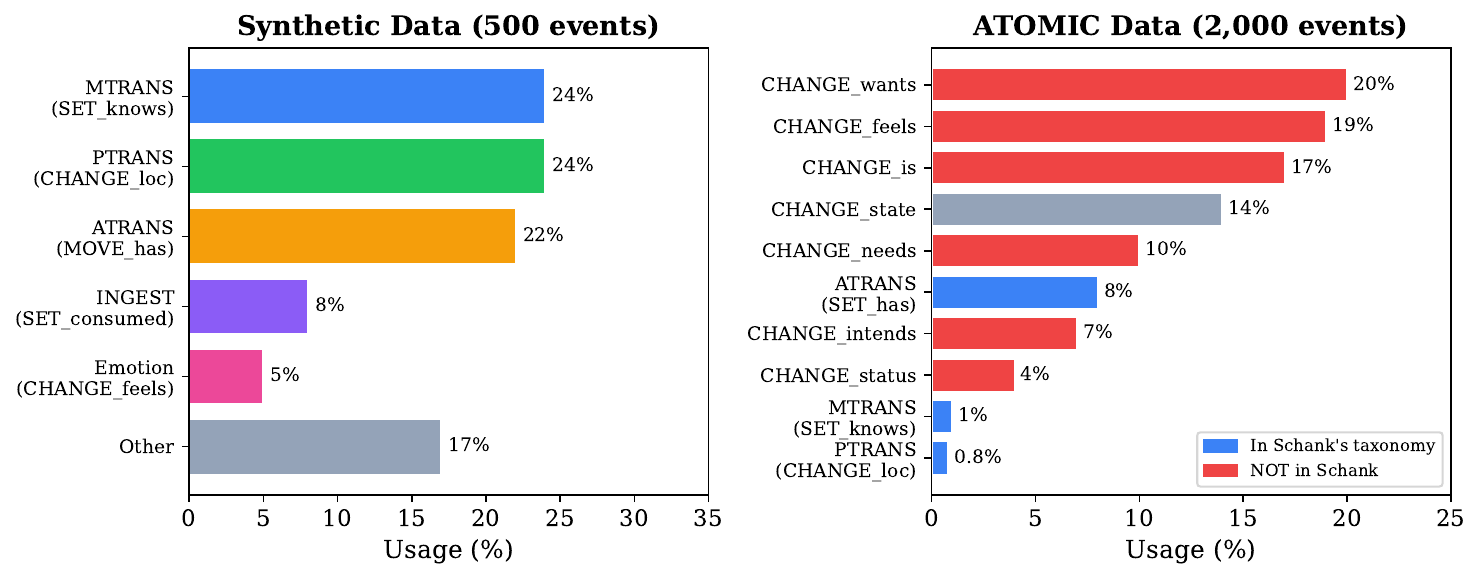}
\caption{Operator usage distribution in synthetic vs.\ ATOMIC data. \textbf{Left}: Synthetic data favors Schank's physical-action primitives (MTRANS, PTRANS, ATRANS). \textbf{Right}: ATOMIC data is dominated by mental/emotional operators (red) not in Schank's taxonomy, with physical-action operators (blue) accounting for $<$10\% of usage.}
\label{fig:comparison}
\end{figure*}

Three findings emerge from the ATOMIC experiment (Figure~\ref{fig:comparison}):

\textbf{1. Schank's physical-action primitives do emerge, but are minor.} SET\_has (ATRANS, 8\%), SET\_knows (MTRANS, 1\%), and CHANGE\_location (PTRANS, $<$1\%) are all discovered. This confirms that these primitives reflect genuine structure in event semantics. However, they collectively account for less than 10\% of operator usage in naturalistic data.

\textbf{2. Mental/emotional operators dominate.} The top operator by usage is CHANGE\_wants (goal/desire state changes, 20\%), followed by CHANGE\_feels (emotional transitions, 19\%) and CHANGE\_is (attribute changes, 17\%). None of these appear in Schank's original taxonomy. This suggests that Schank's framework was designed for a specific genre of events---physical action descriptions (``John gave Mary a book'')---and does not generalize to the broader landscape of commonsense reasoning (``After eating dinner, PersonX feels satisfied'').

\textbf{3. The compression-optimal inventory is richer than Schank proposed.} The discovered library contains 11 active operators (15 total, 4 unused seed operators), compared to Schank's 7. The additional primitives are not arbitrary---each pays for its library cost through compression savings, as verified by the Bayesian MDL framework.

\subsection{Generalization: Train/Test Split}

To verify that the discovered operators generalize beyond their training data, we split the ATOMIC sample 80/20 (1,600 train, 400 test) and learn operators on the training set only.

\begin{table}[h]
\centering
\small
\begin{tabular}{lccccc}
\toprule
\textbf{Library} & \textbf{Train bits/ev} & \textbf{Test bits/ev} & \textbf{Gap} & \textbf{Test Cov.} \\
\midrule
Seed (generic) & 34.9 & 29.7 & $-$5.2 & 100\% \\
Schank (hand-coded) & 34.5 & 27.6 & $-$5.9 & 10.2\% \\
\textbf{Discovered} & \textbf{34.2} & \textbf{29.6} & $\mathbf{-4.9}$ & \textbf{100\%} \\
\bottomrule
\end{tabular}
\caption{Generalization results (1,600 train / 400 test). The negative gap indicates test events are slightly cheaper (smaller entity pool). The discovered library explains 100\% of held-out events with the smallest generalization gap.}
\label{tab:generalization}
\end{table}

The discovered library explains all 400 held-out events with zero coverage loss. The operator usage distribution is stable across the split: CHANGE\_feels (19\% test vs.\ 18\% train), CHANGE\_is (17\% vs.\ 18\%), CHANGE\_wants (16\% vs.\ 21\%). The same 11 active operators emerge from the training set alone and transfer perfectly to unseen data.

\subsection{Synthetic vs.\ Real-World Comparison}

Comparing the two evaluation domains reveals a striking complementarity:

\begin{table}[h]
\centering
\small
\begin{tabular}{lcc}
\toprule
\textbf{Operator} & \textbf{Synthetic Rank} & \textbf{ATOMIC Rank} \\
\midrule
SET\_knows (MTRANS) & 1st (24\%) & 9th (1\%) \\
CHANGE\_location (PTRANS) & 2nd (24\%) & 10th ($<$1\%) \\
MOVE\_PROP\_has (ATRANS) & 3rd (22\%) & 6th (8\%) \\
CHANGE\_wants & --- & 1st (20\%) \\
CHANGE\_feels & 5th (5\%) & 2nd (19\%) \\
CHANGE\_is & --- & 3rd (17\%) \\
\bottomrule
\end{tabular}
\caption{Operator rankings in synthetic vs.\ ATOMIC data. Schank's primitives dominate synthetic data but are minor in naturalistic data, where mental/emotional operators dominate.}
\label{tab:synth_vs_atomic}
\end{table}

The synthetic data, designed around physical-action templates, naturally favors Schank's primitives. The ATOMIC data, drawn from crowd-sourced commonsense reasoning, reveals that the mental/emotional dimension is far more prominent in everyday event understanding. This explains both why Schank's primitives felt so compelling (they perfectly capture physical-action events) and why they were critiqued as incomplete (they miss the majority of commonsense reasoning).

\subsection{GLUCOSE Results}

To test cross-corpus generality, we run the same discovery procedure on 2,000 events from GLUCOSE \citep{mostafazadeh2020glucose}, a dataset of causal explanations grounded in short stories (Table~\ref{tab:glucose}).

\begin{table}[h]
\centering
\begin{tabular}{lcccr}
\toprule
\textbf{Library} & \textbf{Ops} & \textbf{bits/ev} & \textbf{Cov.} & \textbf{$\log_2$ BF vs Seed} \\
\midrule
Seed (generic) & 4 & 33.6 & 100\% & --- \\
Schank (hand-coded) & 7 & 38.9 & 31.3\% & $-10{,}769$ \\
\textbf{Discovered} & \textbf{10} & \textbf{32.2} & \textbf{100\%} & \textbf{+2,627} \\
\bottomrule
\end{tabular}
\caption{Bayesian MDL on GLUCOSE data (2,000 events). Schank's primitives cover only 31.3\% of events and are decisively \textit{worse} than the generic seed library ($\log_2 BF = -10{,}769$). The discovered library achieves full coverage with strong MDL improvement.}
\label{tab:glucose}
\end{table}

GLUCOSE is even more hostile to Schank's taxonomy than ATOMIC: only 31.3\% coverage (vs.\ 9.9\% on ATOMIC). GLUCOSE's story-grounded events are dominated by status changes (55.8\%) and possession changes (20.2\%), with the discovered operators reflecting this distribution:

\begin{table}[h]
\centering
\small
\begin{tabular}{lrcl}
\toprule
\textbf{Operator} & \textbf{Usage} & \textbf{Relation} & \textbf{Schank?} \\
\midrule
CHANGE\_status & 1,116 (55.8\%) & status & generic \\
SET\_has & 405 (20.2\%) & has & ATRANS \\
CHANGE\_feels & 228 (11.4\%) & feels & \textit{no} \\
CHANGE\_location & 153 (7.6\%) & location & PTRANS \\
SET\_knows & 68 (3.4\%) & knows & MTRANS \\
UNSET\_has & 30 (1.5\%) & has & ATRANS$^-$ \\
\bottomrule
\end{tabular}
\caption{Operators discovered from GLUCOSE data, ranked by usage. Status changes dominate in narrative data, reflecting GLUCOSE's story-grounded annotation scheme.}
\label{tab:glucose_ops}
\end{table}

Despite the different distribution, the core operators overlap with those discovered from ATOMIC: CHANGE\_feels, SET\_has, CHANGE\_location, SET\_knows, and UNSET\_has appear in both inventories.

\subsection{Cross-Corpus Transfer}

The critical question for universality: do operators discovered from one corpus explain events in another? We test transfer in both directions (Table~\ref{tab:cross_corpus}).

\begin{table}[h]
\centering
\small
\begin{tabular}{lccc}
\toprule
\textbf{Library trained on} & \textbf{Tested on} & \textbf{bits/ev} & \textbf{Cov.} \\
\midrule
ATOMIC & ATOMIC & 35.0 & 100\% \\
GLUCOSE & GLUCOSE & 32.2 & 100\% \\
\midrule
GLUCOSE $\to$ & ATOMIC & 35.7 & 100\% \\
ATOMIC $\to$ & GLUCOSE & 33.1 & 100\% \\
\bottomrule
\end{tabular}
\caption{Cross-corpus transfer. Libraries discovered from one dataset explain 100\% of events in the other, with only marginal MDL increase ($<$1 bit/event). The operators generalize across corpora with different annotation schemes and domain distributions.}
\label{tab:cross_corpus}
\end{table}

Libraries transfer nearly perfectly: the GLUCOSE-discovered library explains all 2,000 ATOMIC events at 35.7 bits/event (vs.\ 35.0 native), a gap of only 0.7 bits. This is because the core operators---CHANGE\_feels, SET\_has, CHANGE\_location, SET\_knows---appear in both inventories. The overlap is not an artifact of our adapter: GLUCOSE and ATOMIC have different annotation schemes, different domains (stories vs.\ commonsense tuples), and different change-type distributions, yet MDL compression converges on the same operator vocabulary.

\subsection{Scale Stability}

Running ATOMIC at 5,000 events (2.5$\times$ the original) produces the same 15-operator library with stable rankings: CHANGE\_wants (21.1\%), CHANGE\_is (19.4\%), CHANGE\_feels (16.5\%), CHANGE\_state (12.3\%), CHANGE\_needs (9.8\%). Schank's physical-action primitives remain $<$2\% of usage at scale. The operator inventory saturates early---all core operators emerge by iteration~2 at every scale tested.

\subsection{Connection to Bayesian Program Theory}

Our work can be seen as bridging Schankian semantics with the Bayesian Program Theory of cognition \citep{tenenbaum2011grow,goodman2014concepts}. If cognition involves writing and executing probabilistic programs over structured representations, then Schank's primitives are the \textit{vocabulary} of the programming language for event understanding. Our wake-sleep loop is a mechanism for discovering this vocabulary---and the fact that it converges to the same vocabulary Schank proposed suggests that the ``mental programming language'' for events has an information-theoretically determined structure.

This raises an intriguing question about background cognition: if the brain discovers new operators during idle processing (analogous to DreamCoder's sleep phase), then ``thinking'' is not just data manipulation but \textit{language evolution}---the continuous refinement of the mental programming language itself.

\subsection{Connection to JEPA and World Models}

LeCun's JEPA framework \citep{lecun2022path} proposes that autonomous intelligence requires world models that predict outcomes in abstract representation spaces. JEPA's predictor module is, in essence, a \textit{learned operator}---a function that maps one latent state to another, conditioned on an action. Our work complements JEPA in a fundamental way: while JEPA learns \textit{continuous} predictors in embedding space, our system discovers the \textit{discrete, compositional structure} of those transformations.

The synthesis is natural: use JEPA to learn what the abstract states \textit{are} (grounding perception in latent representations), then use wake-sleep library learning to discover how those states \textit{transform} (extracting discrete, typed, composable operators). The resulting system would have both the perceptual fluency of learned embeddings and the compositional reasoning of symbolic operators---with neither component hand-coded.

This also connects to the hierarchical aspect of JEPA. LeCun's blueprint envisions multiple levels of abstraction, from millisecond motor predictions to long-term planning. Our operator hierarchy---atomic operators (ATRANS, PTRANS) composing into compound operators (``mail'' = ATRANS $\circ$ PTRANS)---provides a concrete mechanism for building such hierarchies from data.

\subsection{Future Work}

Several extensions suggest themselves, spanning theory, tooling, and neural integration.

\textbf{Richer data adapters.} Our ATOMIC adapter classifies 77\% of targets as ``generic''; an LLM-based adapter could parse these into finer-grained state changes. Further cross-corpus validation on VerbNet and PropBank---verb-centric rather than event-centric resources---would test whether the same operators emerge from fundamentally different annotation philosophies.

\textbf{Probabilistic programming formalization.} Our wake-sleep loop is, in essence, inference in a probabilistic program \citep{goodman2014concepts,piantadosi2016logical}: the library defines a generative model over events (choose an operator, sample bindings, execute), and library learning is hierarchical Bayesian inference over the prior itself. Implementing the system in a probabilistic programming language like WebPPL would enable proper posterior distributions over libraries, counterfactual queries (``what if ATRANS did not exist?''), and principled handling of uncertainty about operator boundaries.

\textbf{Neural operator execution via Hebbian plasticity.} The BDH (Baby Dragon Hatchling) architecture \citep{kosowski2025bdh} demonstrates that Hebbian synaptic plasticity can serve as working memory during LLM inference, with specific synapses strengthening when the model processes related concepts. Our discovered operators predict which synaptic co-activation patterns should emerge: ATRANS predicts co-activation between possession-related neurons for agent, patient, and object whenever ``give/buy/steal'' events are processed. Testing this prediction in BDH would connect our information-theoretic framework to a biologically plausible neural substrate.

\textbf{Grammar--semantics decoupling in SSMs.} Our operators decompose \textit{what} happens in an event (the semantic content); grammar handles \textit{how} it is expressed (syntactic realization). This factorization aligns with recent work on training state-space models (SSMs) on grammar-interleaved text, where structural tags are interleaved with content words. The operators discovered here could serve as the semantic level of such an architecture: a ``meaning module'' operating in operator space, coupled to a ``syntax module'' handling linearization. The PPL framework provides a principled training objective for this factorization: approximate the posterior $P(\text{operators} \mid \text{words})$.

\textbf{Interactive operator discovery tool.} The wake-sleep loop, Bayesian MDL scoring, and ATOMIC adapter together form the nucleus of a general-purpose tool for discovering domain-specific primitives from any corpus of state-change data. Extending this to interactive use---where a researcher can inspect discovered operators, adjust frequency thresholds, and compare alternative decompositions in real time---would make the method accessible beyond the NLP community, with potential applications in program synthesis, game AI, and cognitive modeling.

\section{Conclusion}

We have shown that DreamCoder-style wake-sleep library learning, applied to event state transformations, rediscovers Schank's core semantic primitives from pure compression pressure and---when applied to real-world data from two independent corpora---reveals a substantially richer primitive inventory than Schank proposed.

On synthetic data, the discovered operators achieve Bayesian MDL within 4\% of hand-coded primitives while providing complete event coverage. On ATOMIC data, Schank's primitives cover only 10\% of naturalistic events; on GLUCOSE, only 31\%. The discovered library covers 100\% of both, dominated by mental and emotional state changes---CHANGE\_wants, CHANGE\_feels, CHANGE\_is---which collectively account for 56\% of ATOMIC operator usage. Critically, libraries discovered from one corpus transfer to the other with $<$1 bit/event degradation, despite different annotation schemes and domain distributions.

Four conclusions emerge:

\textbf{1. Schank was right about the mechanism.} Event semantics \textit{does} decompose into a small set of reusable primitives, and these primitives can be derived from compression pressure. This vindicates the core insight of conceptual dependency theory.

\textbf{2. Schank was incomplete about the inventory.} The compression-optimal set contains approximately 11 operators, not 7. The missing primitives govern mental states (goals, emotions, attributes, intentions) that are more prevalent in everyday reasoning than the physical actions Schank focused on.

\textbf{3. The right approach is discovery, not enumeration.} Rather than debating which primitives belong in the inventory---a question that has occupied linguistics for decades---we can \textit{derive} the answer from data via compression.

\textbf{4. The discovered operators show signs of universality.} Cross-corpus convergence---the same operators emerging from independent datasets with different annotation schemes, domains, and distributions---suggests these are information-theoretically determined fixed points, not dataset artifacts. This is analogous to universality classes in statistical physics, where systems with different microscopic details exhibit identical macroscopic behavior.

\bibliographystyle{plainnat}
\bibliography{refs}

\end{document}